\documentclass[letterpaper]{article} 
\usepackage[preprint]{aaai2027}  
\usepackage[hyphens]{url}  
\usepackage{graphicx} 
\usepackage{natbib}  
\usepackage{caption} 
\usepackage{amsmath}
\usepackage{amssymb}
\usepackage{booktabs}
\usepackage{xcolor}
\usepackage{multirow}
\usepackage{comment} 
\usepackage{adjustbox}
\newcommand{\autoref}[1]{Figure~\ref{#1}}  
\usepackage[ruled,vlined]{algorithm2e} 
\usepackage{tabularx}
\usepackage{array}
\usepackage[table]{xcolor} 

\definecolor{sgpurple}{HTML}{7B2FBE}

\newcommand{\encS}{\mathrm{enc}_S}
\newcommand{\encV}{\mathrm{enc}_V}

\newcommand{\ourmodel}{\textsc{SNIP++}}
\newcommand{\todo}[1]{}
\definecolor{aiedit}{HTML}{0B6E6E}

\newcommand{\sg}[1]{}
\newcommand{\bl}[1]{}
\newcommand{\km}[1]{}

\title{SNIP++: Fine-Grained Symbolic–Numerical Alignment for Symbolic Regression}

\author{Benjamin L\'eger\textsuperscript{\rm 1,2},
Shubham Gupta\textsuperscript{\rm 1,2},
Samy Mammeri\textsuperscript{\rm 1,2},
Kazem Meidani\textsuperscript{\rm 3},
Cem Subakan\textsuperscript{\rm 1,2},
Christian Gagn\'e\textsuperscript{\rm 1,2,4}}
\affiliations{\textsuperscript{\rm 1}Universit\'e Laval\quad
\textsuperscript{\rm 2}Mila\quad
\textsuperscript{\rm 3}Department of Mechanical Engineering, Carnegie Mellon University\quad
\textsuperscript{\rm 4}Canada CIFAR AI Chair\\
\texttt{benjamin.leger.1@ulaval.ca}
}

\begin{document}
\maketitle

\begin{abstract}
%

Mathematical expressions and the numerical behavior they produce are two views of the same underlying function, and connecting them is central to scientific discovery. Symbolic Regression (SR) relies on this connection directly: it searches for an expression that reproduces a given behavior. Recent multi-modal models learn this connection by embedding symbolic expressions and their numerical behavior in a shared representation space. We show that this embedding space is only globally aligned: complete expressions correspond to complete behaviors, but the contribution of individual parts of an expression is not represented. This granularity gap leaves the model unable to tell how a local edit to an expression changes its behavior, the central operation in SR. We introduce a compositional alignment method that closes this gap: a structural positional encoding exposes the substructure of an expression to the encoder, and a multi-granularity contrastive objective grounds each subexpression in the behavior it produces before propagating this grounding to the full expression. 
The resulting
representations close much of the modality gap between symbolic and numerical embeddings, reliably distinguish the effects of local edits that the original alignment cannot, and transfer to external
SR corpora.


\end{abstract}


\section{Introduction}

Symbolic regression (SR) seeks to recover symbolic expressions from numerical observations, connecting two fundamentally different representations of the same function: its mathematical form and its observed behavior \citep{lacava2021srbench}. These two representations live in very different spaces: symbolic expressions are discrete, compositional objects, built from operators and variables according to a tree structure, while numerical behavior is a continuous signal observed at a finite set of points. 

Recent work has approached this through multi-modal learning, most notably SNIP \citep{meidani2024snip}, which embeds symbolic expressions and their numerical behavior in a shared representation space using contrastive learning. Beyond its use within a particular regression pipeline \citep{meidani2024snip, li2026gensr, li2025mmsr, yu2024mdlformer}, such an alignment holds the promise of a general-purpose representation of the correspondence between symbolic structure and numerical behavior. 

However, the relationship between the two is also far from simple: a small change to an expression can leave its behavior essentially unchanged, or alter it completely, so structural similarity in one space does not translate into proximity in the other \citep{krawiec2011learnable}. Also, while mapping a symbolic expression to its numerical behavior is explicit (evaluating it on a fixed set of inputs uniquely determines the corresponding outputs), the difficulty of bridging the symbolic and numerical modalities lies primarily in learning the inverse and relational structure of this correspondence: identifying which symbolic structures are compatible with an observed behavior, and how changes in those structures affect it. This is a central theme in SR \citep{kim2014probabilistic, krawiec2011learnable}. In addition to being poorly localized—with small symbolic alterations potentially inducing large numerical changes—the correspondence is non-unique: distinct symbolic forms can induce the same behavior, especially over a finite set of observations.

Hence, a truly useful shared space must organize symbolic forms by their functional meaning—remaining invariant to equivalent rewritings while sharply distinguishing local, behavior-altering changes \citep{allamanis2017semvec, zheng2025gen, liskowski2020program}. This requires fine-grained, compositional grounding of an expression's internal structure. Because SNIP serializes expressions as prefix token sequences and relies on a global contrastive objective over pooled representations, it captures only coarse cross-modal compatibility. Empirically, it cannot reliably discriminate closely related symbolic variants \citep{leger2026alignment}. We refer to this failure to resolve how fine-grained symbolic structure relates to numerical behavior as the \textit{granularity gap}.

We introduce \textbf{SNIP++}, a compositional alignment method that addresses this granularity gap by grounding the shared representation at the level of an expression's parts. First, a tree-structural positional encoding exposes parent-child relations and subtree membership directly to the symbolic encoder. Second, a multi-granularity contrastive objective grounds each subtree—both contextually within its parent and as a standalone expression—in the specific numerical behavior it produces. This shared target encourages a consistent functional representation across contexts and helps resolve ambiguity in the numerical-to-symbolic direction. Inspired by fine-grained vision–language alignment methods such as FG-CLIP \citep{xie2025fg}, we complement this local supervision with controlled symbolic perturbations that teach the model which structural changes alter behavior. 
Together, these components organize the shared space around functional structure rather than whole-expression similarity, closing SNIP's granularity gap.


\paragraph{Contributions.}
Our main contributions are:
\begin{itemize}
    \item We introduce SNIP++, a fine-grained symbolic--numerical alignment
    method that combines tree-structural positional encoding with
    multi-granularity behavioral supervision.

    \item We show that SNIP++ substantially improves global symbolic--numerical
    alignment and fine-grained local discrimination, while also reducing the
    modality gap between symbolic and numerical representations.

    \item We show that these gains extend beyond the training distribution:
\ourmodel{} substantially outperforms SNIP on held-out edit families and
external, independently curated SR benchmarks, while yielding a more functionally meaningful
representation space.
\end{itemize}

\section{Related Work}
\label{sec:related}

\paragraph{Neural and multimodal symbolic regression.}
Pretrained Transformers have been used to generate symbolic expressions directly from numerical observations, either in a zero-shot manner
\citep{biggio2021nesymres,kamienny2022e2e,vastl2024symformer,li2022transformer} or by planning symbolic search at inference time
\citep{voigt2025generalization,shojaee2023tpsr}.
These approaches primarily encode numerical data and decode symbolic solutions. In contrast, \citet{meidani2024snip}, drawing on CLIP \citep{radford2021clip}, learn a shared representation of symbolic expressions and their numerical behavior, supporting property retrieval and generative symbolic regression when paired with a decoder. Subsequent methods use related multimodal representations primarily for expression generation or latent-space search \citep{li2025mmsr,li2026gensr}.
Our focus is complementary: rather than introducing a decoder or search procedure, we improve the alignment itself so that it captures local symbolic changes and can support candidate comparison, evaluation, and search guidance.

\paragraph{Fine-grained and structure-aware representations.}
Global multimodal contrastive models often struggle with fine-grained semantic changes and compositional relations, and may organize their modalities into separated subspaces
\citep{liang2022modalitygap,wang2023equivariant,lewis2024does,xie2025fg}.
Fine-grained methods address these limitations using local contrasts and token--region alignment
\citep{zhong2022regionclip,jing2024fineclip,xie2025fg}.  In the context of symbolic regression, symbolic-only embedding models learn representations that preserve semantic equivalence despite syntactic variation \citep{allamanis2017semvec,zheng2025gen}. In parallel, structure-aware Transformers introduce positional representations derived from tree paths or node relations, making hierarchical structure
explicit when trees are linearized as token sequences \citep{shiv2019novel, peng2021treepath}.
We combine these directions cross-modally: tree-structural positions expose the compositional organization of an expression, while computational subtrees serve as exact local units whose contextual and standalone representations can be grounded in numerical behaviors evaluated directly.

\paragraph{Locality and compositional search.}
The relationship between symbolic changes and their numerical consequences has long been studied in Genetic Programming through semantic locality, genotype--phenotype mappings, modular representations, and compositional search operators
\citep{kim2014probabilistic,krawiec2013approximating,moraglio2015semantic,krawiec2022compositional}.
These methods seek components that produce desired intermediate behaviors or identify behaviors needed locally to reach a target.
RAG-SR  \citep{zhang2025rag} extends a related component-retrieval perspective to neural symbolic regression.
Our method instead learns a bidirectional symbolic--numerical representation of such components, providing a reusable semantic signal without prescribing a particular search heuristic.

\section{Background and Problem Formulation}
\label{sec:preliminaries}

\subsection{SNIP}
\label{sec:snip-background}

SNIP~\citep{meidani2024snip} learns a shared representation space between
symbolic expressions and the numerical behaviors they generate. Given an
expression $f$ and inputs $X$, let $D_f=(X,f(X))$. A symbolic encoder
$\encS$ and numerical encoder $\encV$ produce
$z^S=\operatorname{pool}(\encS(f))$ and $z^V=\encV(D_f)$. A bidirectional
InfoNCE objective (Eq. \ref{eq:bidirectional-infonce})
aligns matched expression--behavior pairs while separating
mismatched pairs, supporting both symbolic-numerical
($f\!\rightarrow\!y$) and numerical-symbolic
($y\!\rightarrow\!f$) alignment.

On the symbolic side, SNIP serializes each expression tree in prefix order,
adds sequential positional embeddings, and pools the token representations
into a single expression embedding. 
Its dual-encoder formulation exposes both directions of the learned
correspondence directly: from symbolic expressions to numerical behaviors and
from numerical observations to symbolic candidates. These directions support complementary symbolic-regression operations, including comparing candidates through their behavior and
recovering symbolic expressions from numerical observations
\citep{krawiec2011learnable,lacava2021srbench,meidani2024snip}.
SNIP therefore provides a natural setting and controlled backbone  for studying the quality of
symbolic--numerical alignment independently of any particular decoder or
search procedure.

\subsection{Alignment Requirements}
\label{sec:granularity-gap}

A symbolic--numerical representation maps an expression $f$ and its numerical
behavior $D_f=(X,f(X))$ into a shared space. For this representation to support
SR, it should capture functional correspondence at both the
level of complete expressions and the level of their constituent components.

\textbf{Complete-expression correspondence}: the representation of an
expression should agree with the representation of the behavior it generates,
and distinguish it from behaviors produced by other functions. This establishes
the symbolic--numerical correspondence at the scale of the complete expression,
but places no constraint on how its internal structure is represented.

Fine-grained alignment requires three additional properties. First,
\textbf{component grounding}: because subtrees are meaningful compositional
units, each subtree $s$ should be represented as a distinct structural
component, preserving its boundaries and position within the expression, and
associated with the intermediate behavior $s(X)$ that it computes. Second,
\textbf{functional sensitivity}: because symbolic similarity does not reliably
imply behavioral similarity, nearby expressions should separate when a small
edit changes their behavior, while syntactically different expressions should
remain close when they represent the same function. Third,
\textbf{context-dependent contribution}: because the operations above a
subtree transform its output, the same local change can have different effects
depending on where and how it is composed. A representation should therefore
capture not only what a component computes, but also how its behavior
contributes to the complete function.

\subsection{Granularity Gap}
The problem is therefore to learn a shared representation that preserves
complete-expression correspondence while also capturing local and
compositional functional structure. A global alignment objective, as in SNIP,
constrains only pooled complete-expression representations and does not
directly supervise individual subtrees or the consequences of modifying them.
Moreover, because SNIP represents expression trees as prefix sequences with
sequential positional embeddings, parent--child relations and subtree
membership are not explicitly available to the encoder. Consistent with these
limitations, \citet{leger2026alignment} find that SNIP's alignment cannot
reliably distinguish an expression from closely perturbed symbolic
alternatives. We refer to this underdetermination of local functional structure
by global supervision as the \emph{granularity gap}.



\section{Proposed Method}
\label{sec:method}



\begin{figure}[t]
    \centering
    \includegraphics[width=\columnwidth]{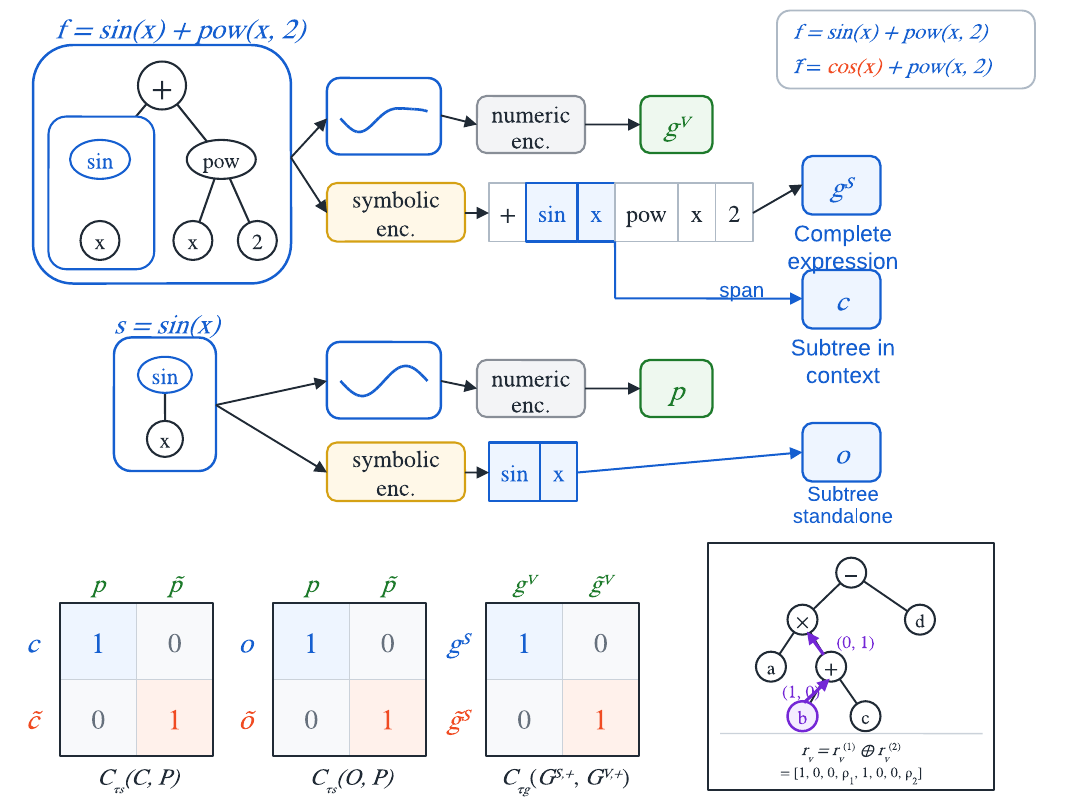}
\caption{\textbf{Overview of SNIP++.}
A controlled edit replaces the subtree \(s=\sin(x)\) with \(\tilde{s}=\cos(x)\), producing the perturbed parent expression \(\tilde{f}\).
From \(f\) and \(s\), the symbolic encoder yields the complete-expression embedding \(g^S\), the subtree span pooled in context \(c\), and the standalone subtree embedding \(o\); \(\tilde{g}^S\), \(\tilde{c}\), and \(\tilde{o}\) denote the corresponding representations of \(\tilde{f}\) and \(\tilde{s}\).
The numerical encoder similarly maps their full-expression and subtree behaviors on shared inputs \(X\) to \(g^V,\tilde{g}^V\) and \(p,\tilde{p}\).
Original and perturbed pairs are aligned symmetrically through the local losses
\(\mathcal{C}_{\tau_s}(C,P)\) and \(\mathcal{C}_{\tau_s}(O,P)\), and the expanded global loss
\(\mathcal{C}_{\tau_g}(G^{S,+},G^{V,+})\), where \(+\) denotes augmenting SNIP's original complete-expression objective with perturbed parent--behavior pairs.
\textbf{Tree-structural position encoding (inset)}: Each token records its nearest \(L\) branch choices (first vs.\ second child)
on the path to the root (\(L=2\) here). In channel \(c\), choice \(k\)
(\(k=0\) nearest) is weighted by learned \(\rho_c^k\), with
\(|\rho_c|<1\); all codes are concatenated so that
\(2CL=d_{\mathrm{model}}\).}
    \label{fig:snippp-method}
\end{figure}

SNIP aligns a pooled symbolic representation of a complete expression with
its numerical behavior, but it neither (a) explicitly represents expression-tree
structure nor (b) supervises the representations of individual subexpressions.
\ourmodel{} addresses these limitations through two complementary modifications. First,
a tree-structural positional encoding makes parent--child relations and subtree
membership available to the symbolic encoder. Second, a multi-granularity
contrastive objective grounds subtrees in the behaviors they compute and
propagates the effects of controlled subtree perturbations to the
representations of complete expressions. The numerical encoder, tokenizer,
and pooling operation remain unchanged. See \autoref{fig:snippp-method} for an overview of the objective function and tree-structural position encoding. 
\subsection{Tree-Structural Positional Encoding}
\label{sec:method-PE}

SNIP linearizes each expression tree in prefix order and adds standard
sequential positional embeddings. These preserve token order but not tree
structure: nodes that are adjacent in the tree may be far apart in the
sequence, while consecutive tokens may belong to different branches. Following
prior tree positional encodings
\citep{shiv2019novel,peng2021treepath}, we therefore augment each token with an
encoding of its location in the expression tree. The inset of
Figure~\ref{fig:snippp-method} illustrates this construction.

For a node \(v\), we record the branch taken at each step toward the root:
\(s_k(v)=(1,0)\) for the first child, \(s_k(v)=(0,1)\) for the second, and
\(s_k(v)=(0,0)\) for padding. Unary operators use the first-child branch. We
retain the nearest \(L\) steps, indexed from the node upward, with \(k=0\)
denoting the nearest relation. In Figure~\ref{fig:snippp-method}, \(b\) is the
first child of \(+\), while \(+\) is the second child of \(\times\), giving
\(s_0(b)=(1,0)\) and \(s_1(b)=(0,1)\). In channel \(c\), these become
\([s_0(b),\rho_c s_1(b)]=[1,0,0,\rho_c]\). With the \(L=2\) and \(C=2\)
channels shown, concatenation yields the eight-dimensional encoding
\(r_b=[1,0,0,\rho_1,\,1,0,0,\rho_2]\).

More generally, channel \(c\) weights relation \(k\) by \(\rho_c^k\), where
\(\rho_c=\tanh(\theta_c)\) is learned. The \(\tanh\) parameterization ensures
\(|\rho_c|<1\), so farther relations decrease in magnitude; different channels
can therefore emphasize different ancestral ranges. Concatenating all
channels gives \(r_v\in\mathbb{R}^{2CL}\), with
\(2CL=d_{\mathrm{model}}\).

For token \(i\), the structural encoding is added to the token and sequential
position embeddings as \(h_i^{(0)}=e_i+q_i+r_{v(i)}\), where \(v(i)\) is its
corresponding tree node. Thus, the encoder retains SNIP's sequential
representation while receiving an explicit description of each token's tree
location. Full construction and initialization details are provided in
Appendix~\ref{app:struct-pe}. This modification changes what the symbolic
encoder can represent but introduces no additional training supervision.

\subsection{Fine-Grained Subtree Alignment}
\label{sec:method-stloss}


Tree-structural positions make subexpressions identifiable to the encoder, but identifiability is not the same as grounding: nothing in SNIP's original objective requires the encoder to use this structure, since supervision is still applied only after pooling the complete expression. We close this gap directly, by supervising sampled subtrees with the numerical behavior they compute. This gives the encoder training signal for exactly the properties identified in Section~\ref{sec:granularity-gap}: component grounding, by tying a subtree's representation to what it computes, and context-dependent contribution, by representing the same subtree both within its parent and on its own.

For each training expression $f_i$ evaluated on inputs $X_i$, we sample a
non-trivial subtree $s_{i0}$ and generate valid perturbations
$s_{i1},\ldots,s_{iK}$. Each perturbation applies one of the training-time edits in
Table~\ref{tab:eval1-perturbations} while leaving the rest of the subtree
unchanged. We write $f_{ik}=f_i[s_{i0}\leftarrow s_{ik}]$ for the parent
expression containing the original or perturbed subtree, and evaluate all
variants on the same inputs $X_i$.

For each $s_{ik}$, we construct a contextual symbolic representation
$c_{ik}=\operatorname{pool}(\encS(f_{ik})[\operatorname{span}(s_{ik})])$, a
standalone symbolic representation
$o_{ik}=\operatorname{pool}(\encS(s_{ik}))$, and a numerical representation
$p_{ik}=\encV(X_i,s_{ik}(X_i))$. Thus, $c_{ik}$ captures the subtree within
its parent, $o_{ik}$ captures it independently, and $p_{ik}$ captures the
behavior it computes (\autoref{fig:snippp-method}).



The local objective aligns each subtree behavior $p_{ik}$ with two symbolic
views: its contextual representation $c_{ik}$ within the parent expression
and its standalone representation $o_{ik}$. Original and perturbed subtrees
are treated symmetrically, so each perturbation is both a hard alternative to
the original and a positive pair for its own numerical behavior. Let $C$, $O$,
and $P$ collect these representations over the batch.
The local objective is
$\mathcal{L}_{\mathrm{local}}
=\mathcal{C}_{\tau_s}(C,P)
+\mathcal{C}_{\tau_s}(O,P)$, 
where $\mathcal{C}_{\tau}$ denotes the
bidirectional contrastive loss and $\tau_s$ is the subtree-level temperature:
\begin{equation}
\label{eq:bidirectional-infonce}
\resizebox{0.90\columnwidth}{!}{$\displaystyle
\mathcal{C}_{\tau}(A,B)
=
-\frac{1}{2M}\sum_{i=1}^{M}
\left[
\log
\frac{\exp(a_i^\top b_i/\tau)}
     {\sum_{j=1}^{M}\exp(a_i^\top b_j/\tau)}
+
\log
\frac{\exp(b_i^\top a_i/\tau)}
     {\sum_{j=1}^{M}\exp(b_i^\top a_j/\tau)}
\right]
$}
\end{equation}
 This grounds each subtree in both what it
computes and how it is represented within its parent. Perturbation and
objective details are provided in
Appendices~\ref{app:perturbations} and~\ref{app:loss-progression}.

\subsection{Complete-Expression Alignment}
\label{sec:method-global-perturb}

Subtree-level supervision constrains the modified component, but does not
directly require the pooled parent representation to capture its effect on the
complete function. For each original or perturbed parent $f_{ik}$, we therefore
form a symbolic representation
$g_{ik}^{S}=\operatorname{pool}(\encS(f_{ik}))$ and a numerical representation
$g_{ik}^{V}=\encV(X_i,f_{ik}(X_i))$.

Let $G^{S}$ and $G^{V}$ collect these complete-expression pairs over the
batch. We define
$\mathcal{L}_{\mathrm{global}}^{+}
=\mathcal{C}_{\tau_g}(G^{S},G^{V})$,
where $\tau_g$ is the complete-expression temperature. The superscript $+$
indicates that SNIP's original global objective is expanded with perturbed
parent-expression pairs, so each local edit also supervises how its numerical
effect is represented after pooling the complete expression (see \autoref{fig:snippp-method}). 

The final objective combines both granularities:
$\mathcal{L}_{\mathrm{full}}
=\mathcal{L}_{\mathrm{global}}^{+}
+\mathcal{L}_{\mathrm{local}}$.
Tree PE makes compositional structure available, the local objective grounds
individual subtrees and their perturbations, and the expanded global objective
propagates these distinctions to complete-expression representations.


\paragraph{Architecture and training.}
\ourmodel{} does not modify SNIP's numerical encoder, tokenizer, or pooling
operation. Both the symbolic and numerical encoders are Transformers with the
same architecture as SNIP's; we refer to \citet{meidani2024snip} for full
architectural detail. Numerical inputs are tokenized using the same
sign--mantissa--exponent encoding as SNIP, adapted from
\citet{charton2021linear}. The only architectural addition is the
tree-structural positional encoding of Section~\ref{sec:method-PE}, which
introduces a small number of additional parameters $\theta_c$ and otherwise
leaves the encoders unchanged. Training proceeds in a single stage: for each
batch, we sample expressions, a subtree and its perturbations per expression,
and compute the contextual, standalone, and complete-expression
representations described above, all produced by the same two encoders.
$\mathcal{L}_{\mathrm{full}}$ is optimized jointly by gradient descent. At
inference, the encoders are used exactly as in SNIP, and all evaluations in
Section~\ref{sec:evaluation} use frozen encoder representations.

\section{Experimental Evaluation}
\label{sec:evaluation}

We evaluate the learned symbolic--numerical alignment directly using frozen
encoder representations and cosine similarity. This isolates the shared
representation from the inductive biases and errors of any particular decoder
or symbolic-search procedure, allowing improvements to be attributed directly
to our method. Because this alignment underlies candidate comparison,
retrieval, search guidance, and downstream generation, its quality provides a
common foundation for a broad range of symbolic regression systems.

Our evaluation proceeds from global correspondence to increasingly
fine-grained functional structure. We first measure
\emph{global symbolic--numerical alignment}, testing whether complete
expressions and their numerical behaviors identify one another among
$100{,}000$ held-out expression--behavior pairs. We then evaluate
\emph{fine-grained symbolic discrimination}, testing whether the shared
embedding space separates closely related symbolic variants according to
their numerical consequences. 
We first test generalization to external expression distributions and use
controlled probes to determine whether the representations track functional
differences while remaining invariant to equivalent symbolic forms. We then use
a cumulative ablation to isolate the contribution of each proposed component.
\subsection{Experimental Setup}
\label{sec:experimental-setup}

\paragraph{Data and comparisons.}
For an expression $f$, we sample inputs $X$ and define its numerical view as
$D_f=(X,f(X))$. Unless stated otherwise, evaluation expressions are sampled
independently from the SNIP generator using random seeds not used during
training, and all symbolic variants of an expression are evaluated on the same
inputs. 
We compare \ourmodel{} with a SNIP
baseline trained under identical conditions, and use the component analysis to
trace how the successive changes from SNIP to \ourmodel{} affect the learned
alignment. Model architecture, optimization, training, and evaluation details
are provided in Appendix~\ref{app:implementation-details}.

\paragraph{Metrics.}
We evaluate global symbolic--numerical alignment in both the
$f\!\rightarrow\!y$ and $y\!\rightarrow\!f$ directions using Recall@10 and
nDCG@10, and measure the modality gap as the Euclidean distance between the
centroids of the $\ell_2$-normalized symbolic and numerical embeddings.
Unless stated otherwise, 95\% confidence intervals are obtained by
bootstrapping evaluation expressions.

\subsection{Global Symbolic--Numerical Alignment}
\label{sec:global-alignment}

We begin by evaluating global correspondence at the level of complete
expressions. For each of the $100{,}000$ held-out expression--behavior pairs
$(f,D_f)$, we use cosine similarity in the shared embedding space to rank all
candidates from the opposite modality. In the $f\!\rightarrow\!y$ direction,
the symbolic representation of an expression must identify its corresponding
numerical behavior; in the $y\!\rightarrow\!f$ direction, the numerical
representation must identify the expression that generated it. The latter
direction is particularly relevant to the inverse problem underlying symbolic
regression, where numerical observations are given and a corresponding
symbolic expression must be identified
~\citep{lacava2021srbench, koza1994genetic}. 

\begin{figure}[t]
    \centering
    \includegraphics[width=0.9\columnwidth]{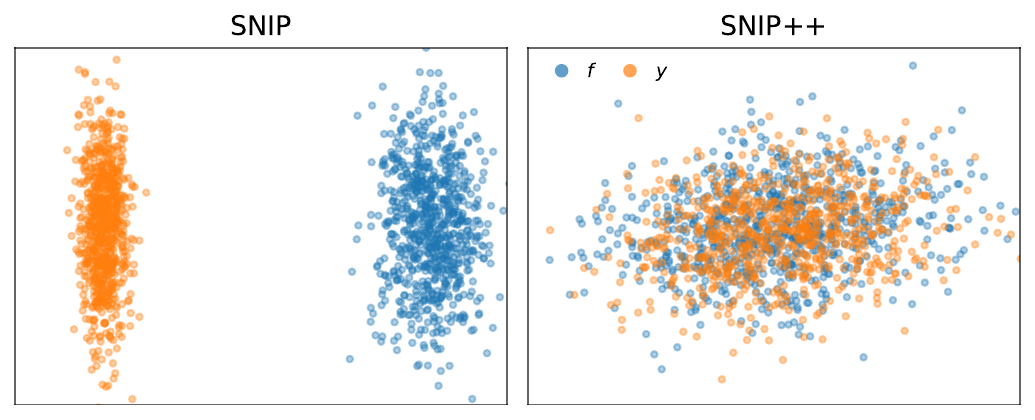}
    \caption{\textbf{Joint symbolic--numerical embedding geometry.}
PCA projections of symbolic ($f$) and numerical ($y$) representations for
$800$ held-out expression--behavior pairs. SNIP separates the two modalities,
whereas \ourmodel{} substantially interleaves them and reduces the modality
gap from approximately $1.2$ to $0.3$.}
    \label{fig:eval1-pca}
\end{figure}

Table~\ref{tab:global-retrieval} shows that \ourmodel{} substantially
improves global symbolic--numerical  alignment in both directions. nDCG@10 increases from
$47.8\%$ to $87.3\%$ for $f\!\rightarrow\!y$ and from $31.3\%$ to
$82.6\%$ for $y\!\rightarrow\!f$, while Recall@10 follows the same trend.
The alignment also becomes more balanced: the nDCG@10 gap
between the two directions decreases from $16.5$ to $4.7$ points. 
Figure~\ref{fig:eval1-pca} shows the same geometric improvement: SNIP
separates symbolic and numerical representations, whereas \ourmodel{}
substantially interleaves them and reduces the centroid distance from
approximately $1.2$ to $0.3$. Together, these results show that \ourmodel{}
establishes a more accurate and balanced correspondence between complete
expressions and their numerical behaviors in the shared embedding space.

\begin{table}[t]
\centering
\small
\setlength{\tabcolsep}{4pt}

\begin{adjustbox}{max width=\columnwidth}
\begin{tabular}{lcccc}
\toprule
& \multicolumn{2}{c}{$f \rightarrow y$}
& \multicolumn{2}{c}{$y \rightarrow f$} \\
\cmidrule(lr){2-3}
\cmidrule(lr){4-5}
Model
& R@10$\uparrow$
& nDCG@10$\uparrow$
& R@10$\uparrow$
& nDCG@10$\uparrow$ \\
\midrule
\textsc{SNIP}
& 70.2
& 47.8
& 53.2
& 31.3 \\
\textbf{\ourmodel{} (ours)}
& \textbf{97.2}
& \textbf{87.3}
& \textbf{94.0}
& \textbf{82.6} \\
\bottomrule
\end{tabular}
\end{adjustbox}

\caption{\textbf{Global symbolic--numerical alignment.}
Recall@10 and nDCG@10 in both alignment directions over $100{,}000$ held-out
expression--behavior pairs. \ourmodel{} substantially improves both directions,
with the largest gains for $y\!\rightarrow\!f$. All values are percentages.}
\label{tab:global-retrieval}
\end{table}

\subsection{Fine-Grained Symbolic Discrimination}
\label{sec:eval1-perturbations}

Global symbolic--numerical alignment tests whether the model can associate an
expression with its numerical behavior among a large set of independently
sampled expressions. It does not, however, establish whether the shared
embedding space can distinguish closely related symbolic variants according
to their numerical consequences. This fine-grained sensitivity is important
in symbolic regression, where candidate expressions often differ by small
symbolic changes that can induce large behavioral differences \cite{krawiec2011learnable, moraglio2015semantic}.


\begin{table}[t]
\centering
\small
\setlength{\tabcolsep}{4pt}
\renewcommand{\arraystretch}{1.04}
\resizebox{0.80\columnwidth}{!}{%
\begin{tabular}{
    @{}
    >{\raggedright\arraybackslash}p{0.42\columnwidth}
    >{\raggedright\arraybackslash}p{0.52\columnwidth}
    @{}
}
\toprule
\textbf{Edit type} & \textbf{Example} \\
\midrule
\rowcolor{black!7}
\multicolumn{2}{@{}l}{\textbf{\textsc{Seen in training}}} \\[1pt]
Unary swap
& $\sin(a)\!\rightarrow\!\cos(a)$ \\
Variable substitution
& $x_0\!\rightarrow\!x_1$ \\
Binary swap
& $a+b\!\rightarrow\!a\times b$ \\
\midrule
\rowcolor{black!7}
\multicolumn{2}{@{}l}{\textbf{\textsc{Held out}}} \\[1pt]
Tree reshaping
& $(a-b)-c\!\rightarrow\!a-(b-c)$ \\
Operand reordering
& $a-b\!\rightarrow\!b-a$ \\
Full-subtree rewrite
& $s\!\rightarrow\!\widetilde{s}$ \\
\bottomrule
\end{tabular}
}
\caption{\textbf{Controlled symbolic edits.}
The first three edit types are used during training; the final three test
generalization to unseen transformations. All evaluation expressions and
individual edits are held out.}
\label{tab:eval1-perturbations}
\end{table}

To test fine-grained alignment, we extend the protocol of \citet{leger2026alignment} and form a local candidate set around each
held-out expression. Starting from an expression $f$, we generate controlled
symbolic variants $\{\widetilde{f}_j\}$ using the six edit types in
Table~\ref{tab:eval1-perturbations}. We then evaluate the original and all
variants on the same inputs $X$, producing the paired behaviors
$f(X)$ and $\{\widetilde{f}_j(X)\}$. The model must distinguish the correct
expression--behavior pairing from the nearby alternatives: in the
$f\!\rightarrow\!y$ direction, $f$ must identify $f(X)$ among the perturbed
behaviors, while in the $y\!\rightarrow\!f$ direction, $f(X)$ must identify
$f$ among the perturbed symbolic forms.

We apply this protocol to 128 held-out expressions, generating valid variants
across all six edit types. Variants that are invalid,
non-finite, or numerically indistinguishable from the original are discarded (details in Appendix~\ref{app:perturbation-construction}). We report the percentile rank of the correct pairing, defined as the fraction
of perturbed alternatives assigned a lower cosine similarity. A value of $1$
indicates perfect discrimination, while $.5$ is expected under random ranking.

\begin{figure}[t]
    \centering
    \includegraphics[
        width=0.7\columnwidth
    ]{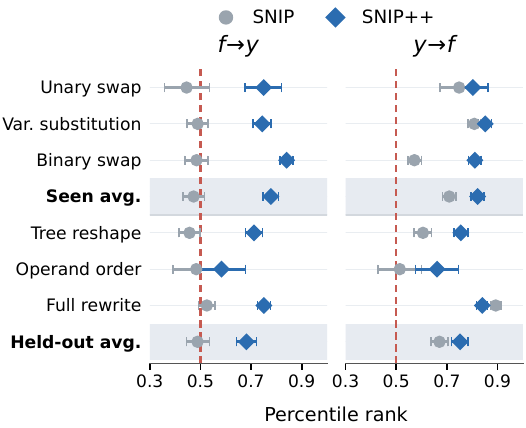}
    \caption{\textbf{Fine-grained symbolic discrimination.}
Percentile rank of the correct complete-expression pairing across six edit
types and both alignment directions. The red dashed line marks chance ($.5$);
shaded rows average the training-represented and held-out edits.}
    \label{fig:eval1-families}
\end{figure}

Figure~\ref{fig:eval1-families} shows that \ourmodel{} more reliably
distinguishes closely related expression--behavior pairs in both alignment
directions. Averaged across edit types, \ourmodel{} reaches percentile ranks
of $.73$ for $f\!\rightarrow\!y$ and $.79$ for $y\!\rightarrow\!f$,
compared with $.48$ and $.69$ for SNIP. Thus, the gains observed in global
alignment carry over to the more fine-grained setting of distinguishing nearby
candidate expressions.
Importantly, the improvement is not confined to edit types represented during
training. It persists across the three held-out transformation types, including
tree reshaping and operand reordering, which modify structural relationships
while preserving the constituent symbols. Overall, \ourmodel{} obtains a
higher point estimate in 11 of the 12 edit-type--direction settings; the sole
reversal occurs for $y\!\rightarrow\!f$ under full-subtree rewriting.
In Appendix~\ref{app:subtree-evaluation}, we also apply the same evaluation to subtrees represented within their parent
expressions, directly testing the local representations targeted by our
objective.



\vspace{-0.1in}
\subsection{Out-of-Domain Generalization}
\label{sec:ood-generalization}


We next test whether the learned alignment transfers beyond expressions
sampled from the training generator. We evaluate on three external
symbolic-regression corpora --- Feynman, Classic, Strogatz \cite{lacava2021srbench,udrescu2020aifeynman,white2013better,strogatz1994nonlinear}, as well as their pooled
union, using two protocols that mirror our main evaluation. First, following
Section~\ref{sec:global-alignment}, we pool all expression--behavior pairs within each corpus and test whether symbolic and numerical queries retrieve their true cross-modal counterparts. Second, we repeat the expression-level fine-grained discrimination analysis from Section~\ref{sec:eval1-perturbations}, where each query is instead ranked only against the small set of controlled perturbations of its own matching expression.

\begin{figure}[t]
    \centering
    \includegraphics[
        width=0.8\columnwidth
    ]{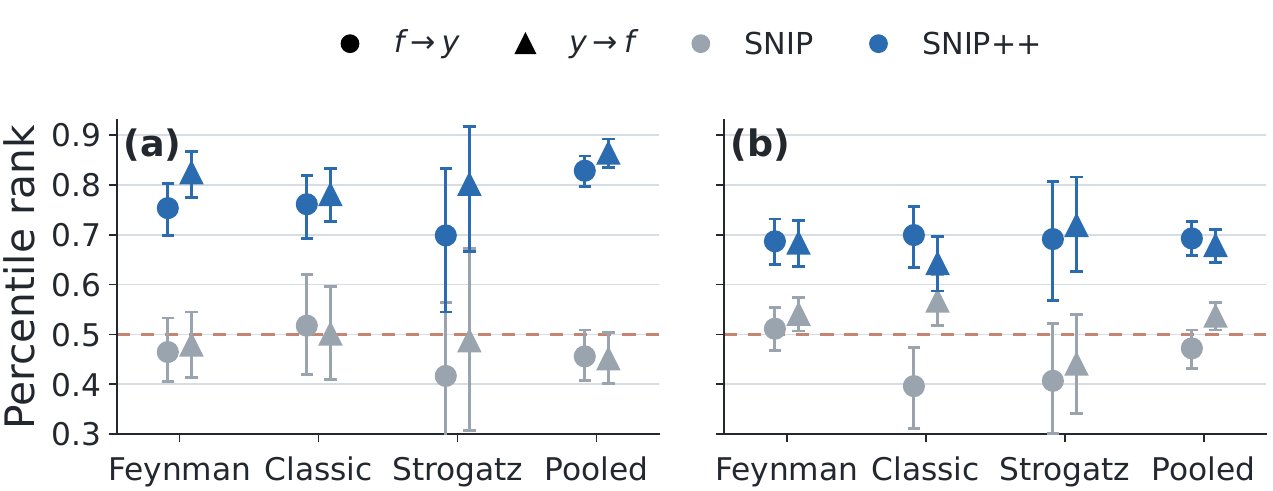}
    \caption{\textbf{Out-of-domain generalization.}
Percentile rank of the correct pairing on Feynman, Classic,
Strogatz and Pooled over all three for (a) within-corpus global symbolic--numerical alignment and
(b) expression-level fine-grained discrimination, evaluated in both
alignment directions. The dashed line marks chance ($.5$).}
    \label{fig:ood-generalization}
\end{figure}

Figure~\ref{fig:ood-generalization} shows consistent transfer under both
protocols. For global, corpus-wide alignment, \ourmodel{} reaches an average percentile
rank of $.79$, compared with $.47$ for SNIP, and remains reliably above
chance in every corpus and direction; SNIP does not in any. For fine-grained, expression-level discrimination, \ourmodel{} reaches $.69$, compared with $.48$, and is
likewise reliably above chance across all corpora, in both directions. Thus, the
improved alignment transfers both globally and locally, to nearby symbolic variants of the same expression.

\begin{figure}[t]
    \centering
    \includegraphics[width=0.9\columnwidth]
        {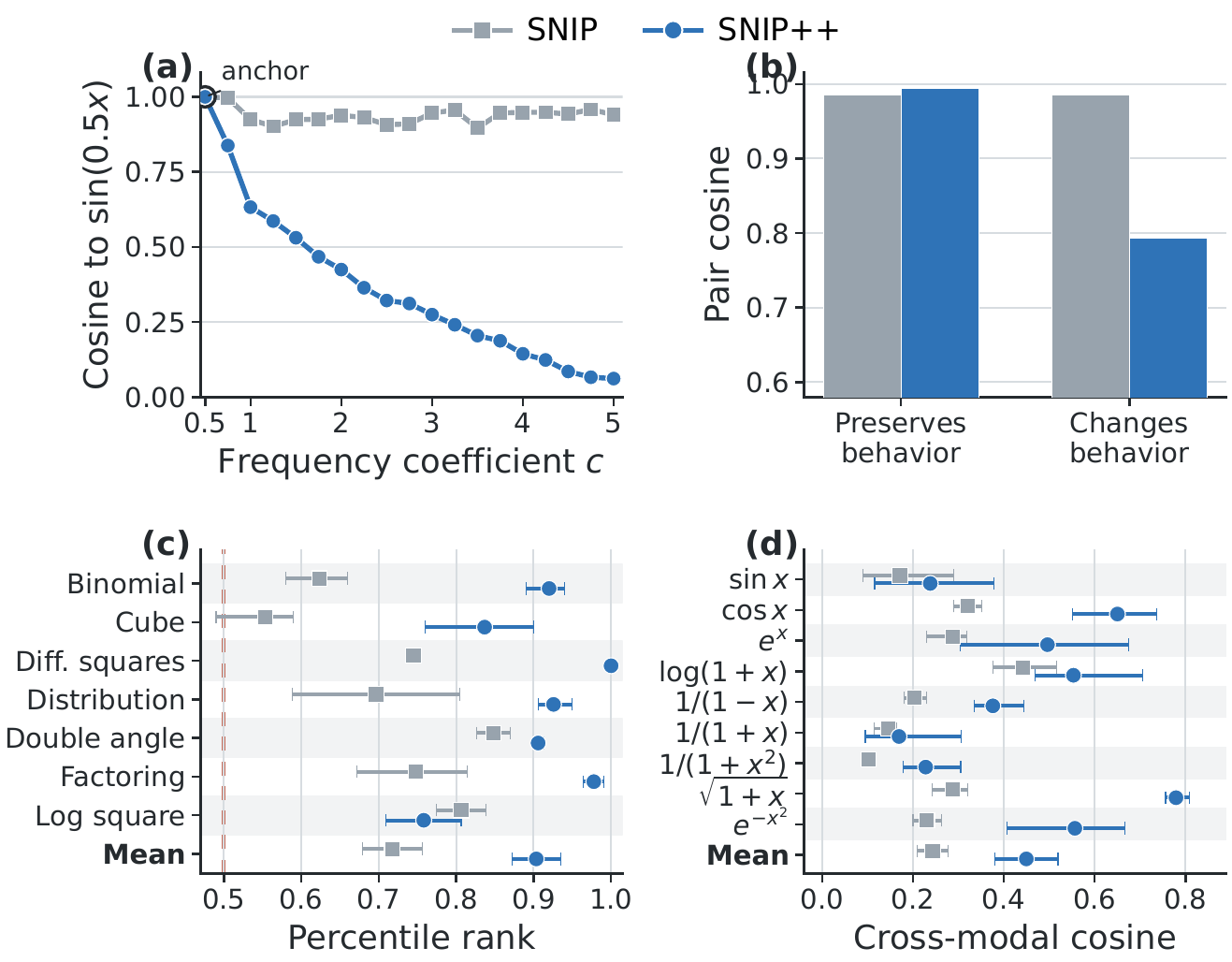}
    \caption{\textbf{Functional structure in the learned space.} (a) continuous functional change, (b) behavior-preserving versus behavior-changing edits, (c) equivalent rewritings versus near-miss alternatives, and (d) cross-modal similarity to Taylor approximations.}
    \label{fig:functional-structure-probes}
\end{figure}

\begin{figure}[t]
    \centering
    \includegraphics[width=\columnwidth]{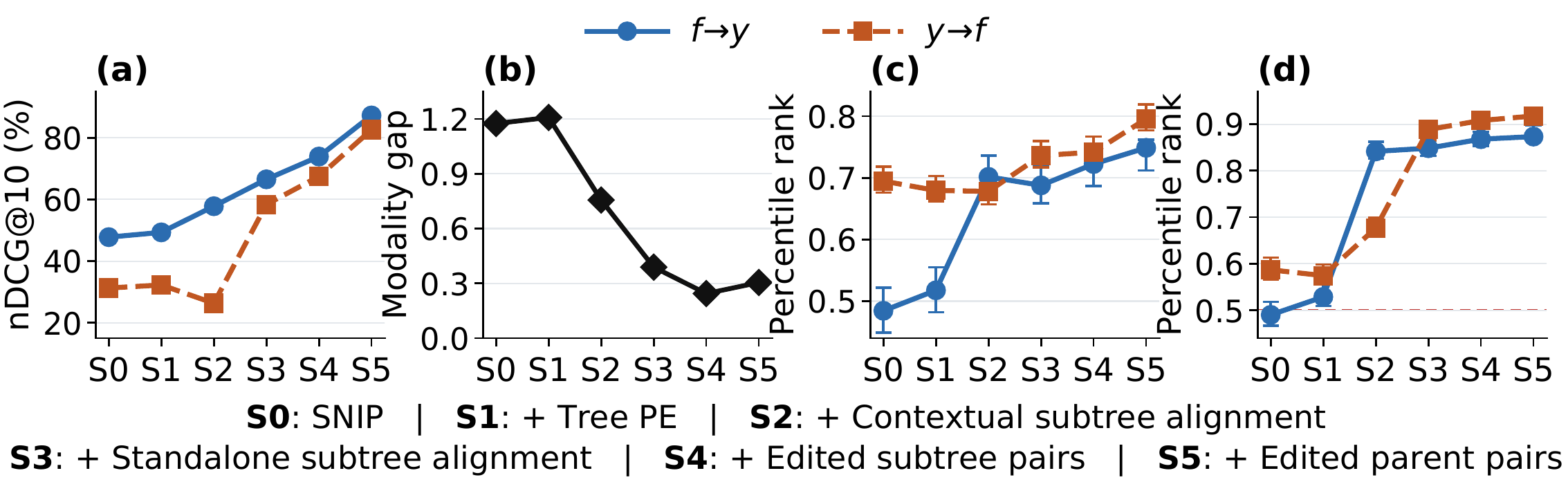}
\caption{\textbf{Cumulative ablation from SNIP to \ourmodel{}.}
(A) global symbolic--numerical alignment, measured by nDCG@10 in both directions; (B) modality gap, measured as the distance between the symbolic and numerical embedding centroids; (C) fine-grained expression-level discrimination; and (D) fine-grained subtree-level discrimination, both measured by percentile rank of the correct pairing. The largest gains come from adding fine-grained subtree supervision, while the later stages further improve balance across directions and recover the full \ourmodel{} performance.}
    \label{fig:abl-ladder}
\end{figure}
\vspace{-0.1in}
\subsection{Functional Structure in the Learned Space}
\label{sec:semantic-geometry}

The preceding evaluations show that \ourmodel{} improves both global
symbolic--numerical alignment and fine-grained discrimination. We next ask
whether these gains produce a functionally meaningful geometry: similar
behavior should induce proximity despite symbolic differences, while similar
symbolic forms should separate when their behaviors diverge.
We examine this geometry through four controlled probes using frozen encoders.
All embeddings are pooled where applicable, \(\ell_2\)-normalized, and compared
using cosine similarity. Figures~\ref{fig:functional-structure-probes}(a)--(c) compare
pairs of symbolic expression embeddings, whereas
Figure~\ref{fig:functional-structure-probes}(d) compares numerical and symbolic
embeddings.
\textbf{Continuous variation} (\autoref{fig:functional-structure-probes}(a)) measures how symbolic similarity changes within
the family $\sin(cx)$ as $c\in[0.5,5]$ moves away from the anchor $c=0.5$.
\textbf{Edit consequences} (\autoref{fig:functional-structure-probes}(b))  compares each expression with an edited version and
groups pairs by whether the edit preserves behavior, as in
\(a+b\rightarrow b+a\), or changes it, as in
\(a-b\rightarrow b-a\). 
\textbf{Equivalent rewritings}(\autoref{fig:functional-structure-probes}(c))
rank a correct algebraic rewrite against approximately $35$ minimally edited
but behaviorally incorrect alternatives across seven identity families.
Finally, \textbf{Symbolic approximations}(\autoref{fig:functional-structure-probes}(d)) measure the cross-modal similarity
between the numerical behavior of nine closed-form functions and their Taylor
truncations of orders $K=2$--$4$.

Figure~\ref{fig:functional-structure-probes} shows a consistent shift toward functional
geometry. Under \textbf{continuous variation}, SNIP still assigns similarity
$.94$ to $\sin(5x)$ and $\sin(0.5x)$, whereas \ourmodel{} decreases smoothly
to $.06$. For \textbf{edit consequences}, SNIP treats behavior-preserving and
behavior-changing edits similarly, while \ourmodel{} keeps the former close
and separates the latter. On \textbf{equivalent rewritings}, \ourmodel{}
raises the mean percentile rank from $.72$ to $.90$ and outperforms SNIP on
six of seven identity families. For \textbf{symbolic approximations},
\ourmodel{} yields higher similarity for all nine functions and increases the
overall mean from $.24$ to $.45$. Together, these results show that
\ourmodel{} separates superficially similar expressions with different
behaviors, while preserving proximity across equivalent and approximate forms.

\vspace{-0.1in}
\subsection{Cumulative Component Analysis}
\label{sec:ablations}

We trace the architectural and objective changes that turn SNIP into
\ourmodel{} and close the granularity gap through six cumulative variants,
each building on the previous one:

\begin{itemize}\setlength\itemsep{2pt}
    \item \textbf{S0:} SNIP, the original baseline.
    \item \textbf{S1:} \textbf{+tree-structural positional encoding}, giving the symbolic encoder the capacity to represent expression structure (Section~\ref{sec:method-PE}), without any new supervision.
    \item \textbf{S2:} \textbf{+contextual subtree alignment}: a subtree representation, pooled from within its parent expression, is aligned with the numerical behavior it computes. Other subtrees' behaviors in the batch serve as easy negatives, and behaviors from training-time perturbations serve as hard negatives.
    \item \textbf{S3:} \textbf{+standalone subtree alignment}: the same numerical behavior is also aligned with the subtree encoded independently of its parent, giving each subtree behavior two symbolic anchors.
    \item \textbf{S4:} \textbf{+symmetric perturbation supervision}: each perturbed subtree is itself encoded symbolically, contextually and standalone, and aligned with its own numerical behavior, rather than serving only as a negative.
    \item \textbf{S5:} \textbf{+complete-expression propagation}: each subtree perturbation is spliced back into its parent expression, and the resulting symbolic--numerical pair is added to the global objective, yielding the full \ourmodel{}.
\end{itemize}

We evaluate each stage using global alignment
(Sec.~\ref{sec:global-alignment}), modality gap, and expression-level
fine-grained discrimination (Sec.~\ref{sec:eval1-perturbations}). We also
repeat the fine-grained analysis at the subtree level, matching subtree spans
pooled from the parent encoding to their numerical behaviors
(App.~\ref{app:subtree-evaluation}).

Figure~\ref{fig:abl-ladder} reveals a clear progression. Tree PE alone changes
little, showing that structural information is not sufficient without
behavioral supervision. Contextual subtree alignment (\textbf{S2}) produces
the largest jump in fine-grained discrimination, especially at the directly
supervised subtree level, but leaves a substantial directional imbalance.
Standalone subtree alignment (\textbf{S3}) restores reverse-direction global
alignment, while symmetric supervision of perturbed subtrees (\textbf{S4})
further balances the two directions. Extending this supervision to perturbed
parent expressions (\textbf{S5}) carries the gains to expression-level
representations. The full \ourmodel{} is best or near-best across all measures
and reduces the modality gap from approximately $1.2$ to $0.3$.

\vspace{-0.07in}
\section{Conclusion and Future Work}


We introduced SNIP++, a fine-grained symbolic--numerical alignment method
combining tree-structural positional encoding with multi-granularity behavioral
supervision. It yields substantial gains in global alignment and fine-grained
discrimination, markedly narrows the modality gap, and transfers to external
SR corpora.

Our ablation clarifies these gains. Exposing expression-tree structure alone is
insufficient; the largest improvements arise when subtrees are aligned with the
behaviors they compute. Encoding each subtree both within its parent and
independently is especially helpful for numerical-to-symbolic alignment, where
multiple symbolic forms can express the same behavior.

This paper focuses on the representation itself rather than embedding SNIP++
inside a complete symbolic-regression system. We therefore leave it to future
work to test how these gains affect end-to-end generation and search. Promising
directions include using the subtree representations to guide symbolic search,
adapting the encoders for generative SR, and developing a reference-free score
for comparing candidate expressions against observed numerical behavior.


\bibliographystyle{aaai2027}
\bibliography{references}
\clearpage
\appendix

\section{PCA Geometry and Variance Spectrum}
\label{app:pca-details}

Figure~\ref{fig:eval1-pca} visualizes the joint symbolic--numerical
geometry of SNIP and \ourmodel{}. For both models, we use the same
$800$ held-out expression--behavior pairs: the first $800$ examples
from the deterministic held-out generator stream used for the
$100{,}000$-pair global retrieval evaluation in
Table~\ref{tab:global-retrieval}. The evaluation set is therefore
identical across models.

For each model separately, we concatenate the $800$ symbolic and
$800$ numerical embeddings into a $1600\times512$ matrix. Each
embedding is individually $\ell_2$-normalized, after which the joint
matrix is mean-centered and PCA is fit using singular-value
decomposition. We apply no feature standardization or whitening, and
all $800$ pairs are displayed in each panel.

For SNIP, the first principal component explains $35.0\%$ of the
variance and the second explains $6.6\%$, for a total of $41.6\%$.
The dominant first component closely follows the separation between
the symbolic and numerical modalities. For \ourmodel{}, the first two
components explain $6.5\%$ and $5.9\%$, respectively, for a total of
$12.4\%$. Their comparable contributions show that modality
separation no longer dominates the leading direction and that the
geometry is distributed more evenly across dimensions.

The complete variance spectrum reinforces this interpretation. SNIP
reaches $50\%$ cumulative variance in four dimensions, reflecting its
heavy leading components, whereas \ourmodel{} uses $11$. The models
require similar dimensionality to capture $90\%$ and $95\%$ of the
variance: $26$ and $35$ dimensions for SNIP, compared with $30$ and
$37$ for \ourmodel{}. At $99\%$, however, \ourmodel{} requires only
$48$ dimensions, compared with $91$ for SNIP. Thus, \ourmodel{} has
both a flatter leading spectrum and a shorter tail: its variance is
less dominated by a modality-separating axis while remaining compact
overall.

The PCA visualization complements the quantitative measurements in
the original embedding space. As reported in
Table~\ref{tab:global-retrieval}, \ourmodel{} substantially improves
bidirectional retrieval, while the distance between the symbolic and
numerical centroids decreases from $1.18$ for SNIP to $0.31$ for
\ourmodel{}.

\section{Tree-Structural Positional Encoding}
\label{app:struct-pe}

This appendix details the tree-structural positional encoding introduced in
Section~\ref{sec:method-PE}.

\subsection{Overview}
\label{app:struct-pe-overview}

SNIP represents an expression using a prefix token sequence and adds a
learned sequential positional embedding to each token. We retain this
representation and augment each symbolic token with a structural vector
derived from the position of its corresponding node in the expression tree.

Let $v(i)$ denote the tree node associated with token $i$. The input to the
symbolic Transformer is
\begin{equation}
    h_i^{(0)}
    =
    e_i + q_i + r_{v(i)},
    \label{eq:app-struct-input}
\end{equation}
where $e_i$ is the token embedding, $q_i$ is SNIP's sequential positional
embedding, and $r_{v(i)}$ is the tree-structural positional encoding.

The structural vector has the same dimension as the token representation:
\begin{equation}
    r_v \in \mathbb{R}^{d_{\mathrm{model}}}.
\end{equation}
We construct it using $C$ channels, each of which represents the nearest
$L$ parent--child relations along the path from node $v$ toward the root.
Each relation is represented by two coordinates, corresponding to the
first and second child slots. Consequently, we choose
\begin{equation}
    2CL=d_{\mathrm{model}}.
    \label{eq:app-struct-width}
\end{equation}

In all experiments, we use $L=64$ and $C=4$ channels.

\subsection{Constructing the Structural Vector}
\label{app:struct-pe-construction}

We construct $r_v$ in three steps: identify the branch relations along the
node-to-root path, weight them within each decay channel, and concatenate
the resulting channel representations.

\paragraph{1. Node-to-root branch relations.}

We number the children of each node according to their argument position.
The first child is assigned slot $1$, the second child is assigned slot
$2$, and the single child of a unary operator is assigned slot $1$.

For a node $v$ at depth $d$, let
\begin{equation}
    a(v)=(a_1,\ldots,a_d),
    \qquad a_j\in\{1,2\},
    \label{eq:app-struct-root-path}
\end{equation}
denote the sequence of child slots encountered when traversing from the
root to $v$. The root has the empty path.

We read this path in the reverse direction, beginning with the relation
between $v$ and its parent. Child slots are represented using the fixed
basis
\begin{equation}
    b(1)=(1,0),
    \qquad
    b(2)=(0,1),
    \qquad
    b(\varnothing)=(0,0),
    \label{eq:app-struct-basis}
\end{equation}
where $\varnothing$ represents padding.

The relation $k$ levels above node $v$ is
\begin{equation}
    s_k(v)=
    \begin{cases}
        b(a_{d-k}), & k<\min(d,L),\\
        b(\varnothing), & \text{otherwise},
    \end{cases}
    \qquad k=0,\ldots,L-1.
    \label{eq:app-struct-relations}
\end{equation}
Thus, $s_0(v)$ describes whether $v$ is the first or second child of its
parent, $s_1(v)$ describes the corresponding relation of its parent, and
so forth. If the path is longer than $L$, only the nearest $L$ relations
are retained. If it is shorter than $L$, the remaining entries are padded
with zero vectors. The root therefore receives only zero relations.

\paragraph{2. Decay within each channel.}

Each channel $c$ has one learned decay parameter
\begin{equation}
    \rho_c=\tanh(\theta_c),
    \qquad \rho_c\in(-1,1).
    \label{eq:app-struct-rho}
\end{equation}
We initialize
\begin{equation}
    \theta_c=\operatorname{artanh}(0.5),
\end{equation}
so that $\rho_c=0.5$ at the beginning of training.

Within channel $c$, the relation $k$ levels above the node is weighted by
$\rho_c^k$. The channel representation is therefore
\begin{equation}
    r_v^{(c)}
    =
    \left[
        s_0(v);
        \rho_c s_1(v);
        \ldots;
        \rho_c^{L-1}s_{L-1}(v)
    \right]
    \in\mathbb{R}^{2L}.
    \label{eq:app-struct-channel}
\end{equation}
The nearest relation receives weight one, while relations farther from the
node receive progressively decayed weights. Different channels can learn
different effective structural ranges: smaller $|\rho_c|$ values emphasize
nearby parent--child relations, while values closer to one preserve
information from more distant ancestors.

\paragraph{3. Channel concatenation.}

The final structural vector is obtained by concatenating the $C$ channel
representations:
\begin{equation}
    r_v
    =
    \left[
        r_v^{(1)};
        \ldots;
        r_v^{(C)}
    \right]
    \in\mathbb{R}^{2CL}
    =
    \mathbb{R}^{d_{\mathrm{model}}}.
    \label{eq:app-struct-full}
\end{equation}
The dimensions are ordered first by channel and then, within each channel,
from the nearest relation to the farthest. Each relation stores the
first-child coordinate followed by the second-child coordinate.

The learned parameters $\{\theta_c\}_{c=1}^{C}$ are the only additional
parameters introduced by Tree PE. No learned projection, gate, or
additional scaling is applied before adding $r_v$ to the token
representation.

\subsection{Token Assignment}
\label{app:struct-pe-token-assignment}

Each token in the prefix serialization is associated with the expression-tree
node it represents and receives that node's structural vector. A node may
produce multiple tokenizer pieces, particularly when representing a
numerical constant. All pieces belonging to the same node receive the same
structural encoding.

Beginning-of-sequence, end-of-sequence, and padding tokens receive the zero
structural vector. Tree PE is applied only to the symbolic encoder; the
numerical encoder, Transformer layers, attention mechanism, normalization,
and pooling operation remain unchanged from SNIP.

\subsection{Worked Example}
\label{app:struct-pe-example}

Consider
\begin{equation}
    f(x_0,x_1)
    =
    \sin\!\left(x_0+2.1\,x_1\right).
\end{equation}
The root-to-node paths and their nearest-first ordering are shown in
Table~\ref{tab:app-struct-example}.

\begin{table}[t]
    \centering

    \begin{minipage}[c]{0.34\columnwidth}
        \centering
        \includegraphics[
            width=\linewidth
        ]{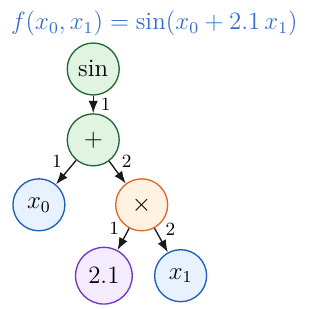}
    \end{minipage}
    \hfill
    \begin{minipage}[c]{0.63\columnwidth}
        \centering
        \scriptsize
        \setlength{\tabcolsep}{2.4pt}
        \renewcommand{\arraystretch}{1.05}

        \begin{tabular}{@{}c c c@{}}
            \toprule
            Node
            & \shortstack{Root-to-node\\path}
            & \shortstack{Node-to-root\\slots} \\
            \midrule
            $\sin$
            & $()$
            & $(\varnothing,\varnothing,\ldots)$ \\

            $+$
            & $(1)$
            & $(1,\varnothing,\ldots)$ \\

            $x_0$
            & $(1,1)$
            & $(1,1,\varnothing,\ldots)$ \\

            $\times$
            & $(1,2)$
            & $(2,1,\varnothing,\ldots)$ \\

            $2.1$
            & $(1,2,1)$
            & $(1,2,1,\varnothing,\ldots)$ \\

            $x_1$
            & $(1,2,2)$
            & $(2,2,1,\varnothing,\ldots)$ \\
            \bottomrule
        \end{tabular}
    \end{minipage}

    \caption{Expression tree and corresponding root-to-node and
    node-to-root child-slot paths for
    $f(x_0,x_1)=\sin(x_0+2.1\,x_1)$.}
    \label{tab:app-struct-example}
\end{table}

For example, $x_1$ has root-to-node path $(1,2,2)$: the addition is the
first child of $\sin$, the multiplication is the second child of the
addition, and $x_1$ is the second child of the multiplication. Reading
these relations from the node toward the root gives $(2,2,1)$ and hence
\begin{equation}
    s_0(x_1)=(0,1),
    \qquad
    s_1(x_1)=(0,1),
    \qquad
    s_2(x_1)=(1,0).
\end{equation}

At initialization, where $\rho_c=0.5$, the representation of $x_1$ in one
channel is
\begin{equation}
    r_{x_1}^{(c)}
    =
    \left[
        (0,1);
        (0,0.5);
        (0.25,0);
        (0,0);
        \ldots
    \right].
    \label{eq:app-struct-example}
\end{equation}
The first pair represents the position of $x_1$ below the multiplication
node, the second represents the position of the multiplication below the
addition, and the third represents the position of the addition below
$\sin$. Concatenating this representation across all $C$ channels produces
the final $d_{\mathrm{model}}$-dimensional vector $r_{x_1}$.

If the constant $2.1$ is split into multiple tokenizer pieces, every piece
receives the structural vector associated with the node path
$(1,2,1)$.

\section{Detailed Contrastive Objectives}
\label{app:loss-progression}

Sections~\ref{sec:method-stloss} and
\ref{sec:method-global-perturb} introduce the subtree-level and
complete-expression objectives used by \ourmodel{}. This appendix defines
the representations and contrastive losses in detail, and then describes
the cumulative objectives evaluated in the component analysis of
Section~\ref{sec:ablations}.

\subsection{Training Examples and Representations}
\label{app:contrastive-representations}

\begin{figure}[t]
    \centering
    \includegraphics[
        width=\columnwidth
    ]{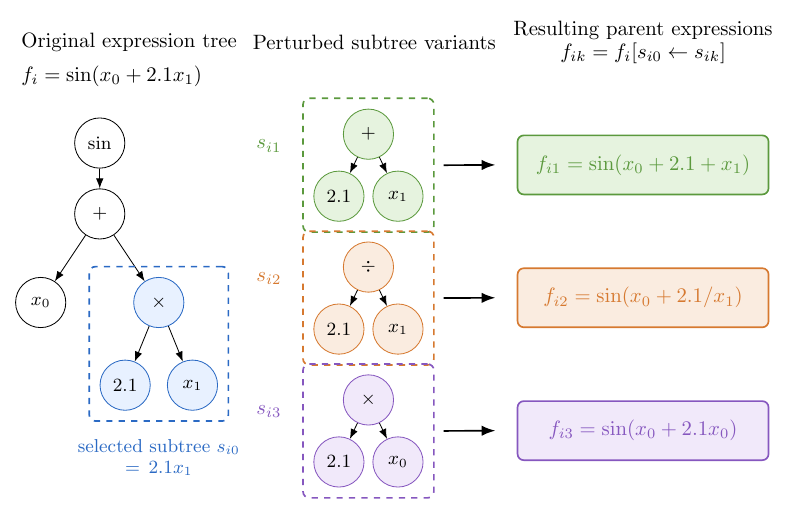}
    \caption{
        Construction of perturbed training examples. Given a parent
        expression $f_i$, we select a subtree $s_{i0}$ and construct
        variants $s_{i1},\ldots,s_{iK_i}$. Each variant is inserted at the
        original subtree location to obtain
        $f_{ik}=f_i[s_{i0}\leftarrow s_{ik}]$. The parent expression and
        all of its variants are evaluated on the same input set $X_i$.
        Colored nodes indicate the modified subtree.
    }
    \label{fig:app-subtree-perturbations}
\end{figure}

For each training expression $f_i$, let $X_i$ denote its sampled input
points. We select an original subtree $s_{i0}$ and construct
$K_i$ valid perturbed variants
$s_{i1},\ldots,s_{iK_i}$. Replacing the original subtree with variant
$s_{ik}$ gives
\begin{equation}
    f_{ik}
    =
    f_i[s_{i0}\leftarrow s_{ik}],
    \qquad
    k=0,\ldots,K_i,
    \label{eq:app-spliced-expression}
\end{equation}
where $f_{i0}=f_i$. The original expression and all of its variants are
evaluated on the same inputs $X_i$. Figure~\ref{fig:app-subtree-perturbations} illustrates this construction:
a subtree $s_{i0}$ is selected from the parent expression $f_i$, modified
to produce variants $s_{ik}$, and spliced back into the same location to
form the corresponding parent expressions $f_{ik}$.

For each original or perturbed subtree, we compute three representations:
\begin{equation}
\begin{aligned}
    c_{ik}
    &=
    \operatorname{pool}\!\left(
        \encS(f_{ik})[\operatorname{span}(s_{ik})]
    \right),\\
    o_{ik}
    &=
    \operatorname{pool}\!\left(
        \encS(s_{ik})
    \right),\\
    p_{ik}
    &=
    \encV\!\left(
        X_i,s_{ik}(X_i)
    \right).
\end{aligned}
\label{eq:app-subtree-representations}
\end{equation}
Here, $c_{ik}$ is the contextual representation of the subtree within its
parent, $o_{ik}$ is its standalone symbolic representation, and $p_{ik}$
represents the numerical behavior computed by the subtree.

We also compute complete-expression representations:
\begin{equation}
\begin{aligned}
    g_{ik}^{S}
    &=
    \operatorname{pool}\!\left(
        \encS(f_{ik})
    \right),\\
    g_{ik}^{V}
    &=
    \encV\!\left(
        X_i,f_{ik}(X_i)
    \right).
\end{aligned}
\label{eq:app-complete-representations}
\end{equation}

Collections marked with a subscript $0$ contain only the original examples ($B$ is the batch size):
\begin{equation}
\begin{aligned}
    C_0 &= \{c_{i0}\}_{i=1}^{B},
    & O_0 &= \{o_{i0}\}_{i=1}^{B},
    & P_0 &= \{p_{i0}\}_{i=1}^{B},\\
    G_0^S &= \{g_{i0}^{S}\}_{i=1}^{B},
    & G_0^V &= \{g_{i0}^{V}\}_{i=1}^{B}.
\end{aligned}
\label{eq:app-original-collections}
\end{equation}
Collections marked with $+$ contain both original and perturbed examples:
\begin{equation}
\begin{aligned}
    C_+ &= \{c_{ik}\}_{i,k},
    & O_+ &= \{o_{ik}\}_{i,k},
    & P_+ &= \{p_{ik}\}_{i,k},\\
    G_+^S &= \{g_{ik}^{S}\}_{i,k},
    & G_+^V &= \{g_{ik}^{V}\}_{i,k}.
\end{aligned}
\label{eq:app-expanded-collections}
\end{equation}
All expanded collections use the same ordering over retained indices
$(i,k)$, so representations with the same index form a matched pair.

\subsection{Contrastive Loss}
\label{app:contrastive-loss}

Let $A=\{a_m\}_{m=1}^{M}$ and $B=\{b_m\}_{m=1}^{M}$ be paired
collections, ordered so that $a_m$ and $b_m$ form a positive pair. We use
the inner product as similarity. The one-direction InfoNCE loss is
\begin{equation}
    \mathcal{I}(A\!\rightarrow\!B;\tau)
    =
    -\frac{1}{M}
    \sum_{m=1}^{M}
    \log
    \frac{
        \exp(a_m^\top b_m/\tau)
    }{
        \sum_{n=1}^{M}
        \exp(a_m^\top b_n/\tau)
    }.
    \label{eq:app-one-direction}
\end{equation}

The bidirectional contrastive loss averages the two retrieval directions:
\begin{equation}
    \mathcal{C}(A,B;\tau)
    =
    \frac{1}{2}
    \left[
        \mathcal{I}(A\!\rightarrow\!B;\tau)
        +
        \mathcal{I}(B\!\rightarrow\!A;\tau)
    \right].
    \label{eq:app-bidirectional}
\end{equation}

\subsection{Complete-Expression and Subtree Objectives}
\label{app:global-local-objectives}

Using the expanded collections, the complete-expression objective is
\begin{equation}
    \mathcal{L}_{\mathrm{global}}^{+}
    =
    \mathcal{C}
    \left(
        G_+^S,G_+^V;\tau_g
    \right).
    \label{eq:app-global-objective}
\end{equation}

The subtree objective aligns each numerical subtree representation with
both of its symbolic views:
\begin{equation}
    \mathcal{L}_{\mathrm{local}}
    =
    \mathcal{C}
    \left(
        C_+,P_+;\tau_s
    \right)
    +
    \mathcal{C}
    \left(
        O_+,P_+;\tau_s
    \right).
    \label{eq:app-local-objective}
\end{equation}

The full objective is
\begin{equation}
    \boxed{
    \mathcal{L}_{\mathrm{\ourmodel{}}}
    =
    \mathcal{L}_{\mathrm{global}}^{+}
    +
    \mathcal{L}_{\mathrm{local}}
    }.
    \label{eq:app-full-objective}
\end{equation}
We use temperature $\tau_g=1$ for complete-expression alignment and $\tau_s=0.1$ for subtree alignment.

\subsection{Cumulative Objectives}
\label{app:cumulative-objectives}

The component analysis in Section~\ref{sec:ablations} evaluates six
cumulative variants. Each stage retains the architecture and supervision
introduced by the preceding stages. We denote the global
complete-expression objective over the original training pairs by
\begin{equation}
    \mathcal{L}_{\mathrm{global}}^{0}
    =
    \mathcal{C}
    \left(
        G_0^S,G_0^V;\tau_g
    \right),
    \label{eq:app-global-original}
\end{equation}
where $\tau_g=1$.

\paragraph{S0: SNIP.}

The baseline uses SNIP's original complete-expression contrastive
objective:
\begin{equation}
    \mathcal{L}^{\mathrm{S0}}
    =
    \mathcal{L}_{\mathrm{global}}^{0}.
    \label{eq:app-stage-s0}
\end{equation}

\paragraph{S1: Tree-structural positional encoding.}

S1 augments the symbolic encoder with the tree-structural positional
encoding of Appendix~\ref{app:struct-pe}. The objective is unchanged:
\begin{equation}
    \mathcal{L}^{\mathrm{S1}}
    =
    \mathcal{L}_{\mathrm{global}}^{0}.
    \label{eq:app-stage-s1}
\end{equation}
The distinction from S0 is therefore architectural rather than
objective-level.

\paragraph{S2: Contextual subtree alignment.}

For each original contextual subtree representation $c_{i0}$, the matching
numerical behavior $p_{i0}$ is the positive example. Numerical behaviors
from other expressions in the batch act as in-batch negatives, while the
behaviors of perturbed subtrees from the same parent act as hard negatives.

For each contextual anchor $c_{i0}$, define the normalizing term
\begin{equation}
\begin{aligned}
    Z^{\mathrm{ctx}}_i
    &=
    \sum_{j=1}^{B}
    \exp\!\left(
        c_{i0}^{\top}p_{j0}/\tau_s
    \right) \\
    &\quad+
    \sum_{k=1}^{K_i}
    \exp\!\left(
        c_{i0}^{\top}p_{ik}/\tau_s
    \right).
\end{aligned}
\label{eq:app-contextual-normalizer}
\end{equation}

The symbolic-to-numerical contextual loss is then
\begin{equation}
    \mathcal{I}_{\mathrm{ctx}}^{S\rightarrow V}
    =
    -\frac{1}{B}
    \sum_{i=1}^{B}
    \log
    \frac{
        \exp\!\left(
            c_{i0}^{\top}p_{i0}/\tau_s
        \right)
    }{
        Z^{\mathrm{ctx}}_i
    }.
\label{eq:app-contextual-forward}
\end{equation}

At this stage, the perturbed subtrees are used only as numerical hard
negatives and are not yet encoded as matched symbolic examples. The
numerical-to-symbolic direction therefore uses only the original pairs:
\begin{equation}
    \mathcal{I}_{\mathrm{ctx}}^{V\rightarrow S}
    =
    -\frac{1}{B}
    \sum_{i=1}^{B}
    \log
    \frac{
        \exp\!\left(p_{i0}^{\top}c_{i0}/\tau_s\right)
    }{
        \displaystyle
        \sum_{j=1}^{B}
        \exp\!\left(p_{i0}^{\top}c_{j0}/\tau_s\right)
    } .
    \label{eq:app-contextual-reverse}
\end{equation}

The bidirectional contextual subtree loss is
\begin{equation}
    \mathcal{L}_{\mathrm{ctx}}^{0}
    =
    \frac{1}{2}
    \left[
        \mathcal{I}_{\mathrm{ctx}}^{S\rightarrow V}
        +
        \mathcal{I}_{\mathrm{ctx}}^{V\rightarrow S}
    \right].
    \label{eq:app-contextual-original}
\end{equation}

The S2 objective is
\begin{equation}
    \mathcal{L}^{\mathrm{S2}}
    =
    \mathcal{L}_{\mathrm{global}}^{0}
    +
    \mathcal{L}_{\mathrm{ctx}}^{0}.
    \label{eq:app-stage-s2}
\end{equation}

\paragraph{S3: Standalone subtree alignment.}

S3 additionally encodes each original subtree independently of its parent.
The standalone loss mirrors the contextual loss from S2, replacing the
contextual symbolic representation $c_{i0}$ with the standalone
representation $o_{i0}$.

For each contextual anchor $c_{i0}$, define the normalizing term
\begin{equation}
\begin{aligned}
    Z^{\mathrm{stand}}_i
    &=
    \sum_{j=1}^{B}
    \exp\!\left(
        o_{i0}^{\top}p_{j0}/\tau_s
    \right) \\
    &\quad+
    \sum_{k=1}^{K_i}
    \exp\!\left(
        o_{i0}^{\top}p_{ik}/\tau_s
    \right),
\end{aligned}
\end{equation}

The symbolic-to-numerical contextual loss is then
\begin{equation}
    \mathcal{I}_{\mathrm{stand}}^{S\rightarrow V}
    =
    -\frac{1}{B}
    \sum_{i=1}^{B}
    \log
    \frac{
        \exp\!\left(
            o_{i0}^{\top}p_{i0}/\tau_s
        \right)
    }{
        Z^{\mathrm{stand}}_i
    }.
\end{equation}

The numerical-to-symbolic term uses the original matched pairs:
\begin{equation}
    \mathcal{I}_{\mathrm{stand}}^{V\rightarrow S}
    =
    -\frac{1}{B}
    \sum_{i=1}^{B}
    \log
    \frac{
        \exp\!\left(p_{i0}^{\top}o_{i0}/\tau_s\right)
    }{
        \displaystyle
        \sum_{j=1}^{B}
        \exp\!\left(p_{i0}^{\top}o_{j0}/\tau_s\right)
    } .
    \label{eq:app-standalone-reverse}
\end{equation}

The bidirectional standalone loss is
\begin{equation}
    \mathcal{L}_{\mathrm{stand}}^{0}
    =
    \frac{1}{2}
    \left[
        \mathcal{I}_{\mathrm{stand}}^{S\rightarrow V}
        +
        \mathcal{I}_{\mathrm{stand}}^{V\rightarrow S}
    \right].
    \label{eq:app-standalone-original}
\end{equation}

The cumulative S3 objective is
\begin{equation}
    \mathcal{L}^{\mathrm{S3}}
    =
    \mathcal{L}_{\mathrm{global}}^{0}
    +
    \mathcal{L}_{\mathrm{ctx}}^{0}
    +
    \mathcal{L}_{\mathrm{stand}}^{0}.
    \label{eq:app-stage-s3}
\end{equation}

\paragraph{S4: Symmetric perturbed-subtree alignment.}

S4 additionally encodes every retained perturbed subtree symbolically.
For each original or perturbed index $(i,k)$, the contextual representation
$c_{ik}$ and standalone representation $o_{ik}$ are paired with the
corresponding numerical behavior $p_{ik}$. Thus, each perturbation becomes
a positive example in both retrieval directions, rather than appearing
only as a numerical hard negative.

Let
\begin{equation}
\begin{aligned}
    C_+ &= \{c_{ik}\}_{i,k},\\
    O_+ &= \{o_{ik}\}_{i,k},\\
    P_+ &= \{p_{ik}\}_{i,k},
\end{aligned}
\label{eq:app-s4-expanded-collections}
\end{equation}
where the three collections use the same ordering over all retained
original and perturbed indices $(i,k)$. Representations with the same
index therefore form a positive pair.

The expanded local objective is
\begin{equation}
    \mathcal{L}_{\mathrm{local}}
    =
    \mathcal{C}(C_+,P_+;\tau_s)
    +
    \mathcal{C}(O_+,P_+;\tau_s),
    \qquad \tau_s=0.1.
    \label{eq:app-expanded-local}
\end{equation}

The S4 objective is
\begin{equation}
    \mathcal{L}^{\mathrm{S4}}
    =
    \mathcal{L}_{\mathrm{global}}^{0}
    +
    \mathcal{L}_{\mathrm{local}}.
    \label{eq:app-stage-s4}
\end{equation}

Equation~\eqref{eq:app-expanded-local} replaces the original-only
contextual and standalone losses used in S3. It does not add a second copy
of those losses: the original pairs $(i,0)$ are already included in
$C_+$, $O_+$, and $P_+$.

\paragraph{S5: Perturbed complete-expression alignment.}

Finally, each original or perturbed subtree $s_{ik}$ is spliced into its
parent expression to form $f_{ik}$. We compute the corresponding
complete-expression representations
\begin{equation}
\begin{aligned}
    g_{ik}^{S}
    &=
    \operatorname{pool}\!\left(
        \encS(f_{ik})
    \right),\\
    g_{ik}^{V}
    &=
    \encV\!\left(
        X_i,f_{ik}(X_i)
    \right).
\end{aligned}
\label{eq:app-s5-global-representations}
\end{equation}

Let
\begin{equation}
    G_+^S=\{g_{ik}^{S}\}_{i,k},
    \qquad
    G_+^V=\{g_{ik}^{V}\}_{i,k},
\end{equation}
where both collections use the same ordering over all retained original and
perturbed indices $(i,k)$. Representations with the same index therefore
form a positive complete-expression pair.

The expanded global objective is
\begin{equation}
    \mathcal{L}_{\mathrm{global}}^{+}
    =
    \mathcal{C}
    \left(
        G_+^S,G_+^V;\tau_g
    \right),
    \qquad
    \tau_g=1.
    \label{eq:app-expanded-global}
\end{equation}

The full \ourmodel{} objective is
\begin{equation}
    \mathcal{L}^{\mathrm{S5}}
    =
    \mathcal{L}_{\mathrm{global}}^{+}
    +
    \mathcal{L}_{\mathrm{local}}.
    \label{eq:app-stage-s5}
\end{equation}

Here, $\mathcal{L}_{\mathrm{local}}$ is the expanded contextual and
standalone subtree objective introduced in S4.

\section{Controlled-Edit Evaluation Details}
\label{app:perturbation-construction}

This appendix provides the construction and filtering details for the
fine-grained complete-expression evaluation in
Section~\ref{sec:eval1-perturbations}.

\subsection{Targets and Input Samples}

We sample $128$ target expressions from a held-out stream of the SNIP
expression generator. Each target expression $f_i$ is associated with its
own input set $X_i$ of $200$ sampled points. The original expression and
all variants derived from it are evaluated on the same $X_i$:
\begin{equation}
    y_i=f_i(X_i),
    \qquad
    \widetilde{y}_{ij}
    =
    \widetilde{f}_{ij}(X_i).
\end{equation}
Thus, differences between $y_i$ and $\widetilde{y}_{ij}$ arise from the
symbolic edit rather than from different input samples.

An expression is retained as an evaluation target only when its output is
finite at every point in $X_i$ and has variance of at least $10^{-10}$.
After filtering, $127$ complete-expression targets remain.

\subsection{Controlled Edit Families}

For each target expression, we construct variants using the six edit
families in Table~\ref{tab:eval1-perturbations}. The first three families
overlap with edits used during training, while the final three are held
out. All target expressions and individual edit instances remain unseen
during training.

\paragraph{Unary operator swap.}

One supported unary operator is replaced by another operator from the same
group:
\[
    \{\sin,\cos,\tan\},\qquad
    \{(\cdot)^2,(\cdot)^3,\sqrt{\cdot}\},\qquad
    \{\exp,\log\}.
\]
This edit is unavailable when the expression contains no supported unary
operator.

\paragraph{Variable substitution.}

One variable is replaced consistently by a different variable already
used in the expression. All occurrences of the selected source variable
are replaced. This edit is unavailable when fewer than two distinct
variables occur.

\paragraph{Binary operator swap.}

One occurrence of an operator in
\[
    \{+,-,\times,\div\}
\]
is replaced by one of the other three operators, while its operands and
the remainder of the tree remain unchanged.

\paragraph{Tree reshaping.}

A local binary chain is re-associated by a tree rotation. For example,
\begin{equation}
    \operatorname{op}
    \bigl(
        \operatorname{op}'(a,b),c
    \bigr)
    \longrightarrow
    \operatorname{op}'
    \bigl(
        a,\operatorname{op}(b,c)
    \bigr).
\end{equation}
The symbols are preserved, but their tree relationships change. This edit
is unavailable when no binary node has a binary child.

\paragraph{Operand reordering.}

The two operands of a subtraction or division node are exchanged:
\[
    a-b\longrightarrow b-a,
    \qquad
    a/b\longrightarrow b/a.
\]
This edit is unavailable when the expression contains neither subtraction
nor division.

\paragraph{Full-subtree rewriting.}

A node is selected and the complete subtree rooted at that node is replaced
by a newly generated subtree. The replacement is sampled from the same
symbolic grammar as the training expressions and contains at most two
binary operators. The root may be selected, in which case the entire
expression is rewritten.

\subsection{Variant Generation and Filtering}
\label{app:eval-filtering}

For each target expression $f_i$ and each applicable edit family, we
generate up to $64$ candidate variants. An edit family is skipped for a
target when its required structure is absent; for example, operand
reordering requires a subtraction or division node.

Every candidate variant $\widetilde{f}_{ij}$ is evaluated on the same input
set $X_i$ as its original expression:
\begin{equation}
    y_i=f_i(X_i),
    \qquad
    \widetilde{y}_{ij}
    =
    \widetilde{f}_{ij}(X_i).
\end{equation}
Using the same inputs ensures that any numerical difference is caused by
the symbolic edit.

A candidate is retained only if it passes the following checks:

\begin{enumerate}
    \item \textbf{Valid expression.}
    The edit must produce a well-formed expression tree that can be
    serialized and evaluated.

    \item \textbf{Finite behavior.}
    Every value in $\widetilde{y}_{ij}$ must be finite. We discard a
    candidate if it produces any \texttt{NaN} or infinite output on $X_i$.

    \item \textbf{Behavior differs from the original.}
    We discard the candidate when
    $\widetilde{y}_{ij}$ is identical to the original output $y_i$.

    \item \textbf{Behavior differs from previously retained variants.}
    We also discard the candidate when its output is identical to that of
    an already retained variant from the same target and edit family.
\end{enumerate}

For the final two checks, output vectors are converted to
\texttt{float32} and compared exactly. Thus, two expressions are called
\emph{numerically indistinguishable} when they produce bit-identical
\texttt{float32} output vectors on the shared $200$ input points.
All candidates that pass these checks are retained. Consequently, the
number of retained variants may differ across targets and edit families.
A target contributes to a family-specific result only when at least one
variant from that family is retained.

\subsection{Evaluation-Pool Statistics}

Table~\ref{tab:perturbation-statistics} summarizes the resulting
complete-expression evaluation pool. A target contributes to a family only
when at least one valid variant of that family is retained.

\begin{table}[t]
    \centering
    \small
    \setlength{\tabcolsep}{4pt}
    \begin{tabular}{lrrr}
        \toprule
        \textbf{Edit family}
        & \textbf{Targets}
        & \textbf{Mean}
        & \textbf{Median} \\
        \midrule
        Unary swap
        & 81
        & 2.1
        & 2 \\
        Variable substitution
        & 119
        & 26.5
        & 27 \\
        Binary swap
        & 127
        & 8.5
        & 9 \\
        Tree reshaping
        & 127
        & 10.7
        & 11 \\
        Operand reordering
        & 90
        & 1.8
        & 2 \\
        Full-subtree rewriting
        & 127
        & 57.6
        & 59 \\
        \bottomrule
    \end{tabular}
    \caption{\textbf{Complete-expression controlled-edit pool.}
    Targets denotes the number of held-out expressions with at least one
    retained variant. Mean and median give the number of retained variants
    per contributing target.}
    \label{tab:perturbation-statistics}
\end{table}

\subsection{Percentile Rank and Aggregation}

All representations are $\ell_2$-normalized for evaluation, and similarity
is measured using cosine similarity. For one target, edit family, and
alignment direction, let $s_i^{+}$ denote the similarity of the correct
expression--behavior pair and let
$s_{i1}^{-},\ldots,s_{iV_i}^{-}$ denote the similarities to its retained
edited alternatives. We define
\begin{equation}
    \operatorname{PR}_i
    =
    \frac{1}{V_i}
    \sum_{j=1}^{V_i}
    \mathbb{I}
    \left[
        s_i^{+}>s_{ij}^{-}
    \right].
    \label{eq:app-percentile-rank}
\end{equation}
A score of $1$ means that the correct pairing outranks every edited
alternative, while $.5$ is expected under random pairwise ranking. We use
a strict inequality, so ties do not count as successful comparisons.

In the $f\!\rightarrow\!y$ direction, the original symbolic expression is
the query and ranks its true behavior against the behaviors of its edited
variants. In the $y\!\rightarrow\!f$ direction, the original numerical
behavior ranks its true symbolic expression against the edited symbolic
variants.

For each family and direction, we average
$\operatorname{PR}_i$ uniformly over contributing target expressions.
Averages over the three training-represented or three held-out families
are computed as means of their family-level scores. Confidence intervals are obtained from $1{,}000$ bootstrap resamples of
the target expressions.

\section{Contextual-Subtree Discrimination}
\label{app:subtree-evaluation}

Section~\ref{sec:eval1-perturbations} evaluates whether complete-expression
representations distinguish an expression from closely related symbolic
variants. Here, we repeat the same evaluation at the subtree level to test
the contextual representations directly targeted by our local training
objective.

\paragraph{Protocol.}

For each of the $128$ held-out parent expressions $f_i$, we select up to
eight non-trivial subtrees. A non-trivial subtree is a strict descendant of
the root that is not a single leaf. When more than eight are available, we
sample eight without replacement using the fixed evaluation seed.

Let $s_{i0}$ denote a selected original subtree and let
$s_{ij}$, $j=1,\ldots,V_i$, denote its retained edited variants. We generate
the variants using the same six edit families and validity filters described
in Appendix~\ref{app:perturbation-construction}. Each edited subtree is
inserted at the original tree position:
\begin{equation}
    f_{ij}
    =
    f_i
    \left[
        s_{i0}\leftarrow s_{ij}
    \right],
    \qquad
    f_{i0}=f_i.
    \label{eq:app-contextual-parent}
\end{equation}

The original subtree and all of its variants are evaluated on the same input
set $X_i$. For each index $j$, we construct a contextual symbolic
representation and a numerical representation:
\begin{equation}
\begin{aligned}
    c_{ij}
    &=
    \operatorname{pool}
    \left(
        \encS(f_{ij})
        [\operatorname{span}(s_{ij})]
    \right),\\
    p_{ij}
    &=
    \encV
    \left(
        X_i,s_{ij}(X_i)
    \right).
\end{aligned}
\label{eq:app-contextual-eval-representations}
\end{equation}
Thus, every symbolic candidate is represented at the same location within
the same surrounding parent structure. Only the selected subtree changes.

\paragraph{Scoring and aggregation.}

In the $f\!\rightarrow\!y$ direction, the original contextual representation
$c_{i0}$ ranks its matching behavior $p_{i0}$ against the behaviors
$\{p_{ij}\}_{j=1}^{V_i}$ of the edited subtrees. In the
$y\!\rightarrow\!f$ direction, $p_{i0}$ ranks $c_{i0}$ against the
contextual representations $\{c_{ij}\}_{j=1}^{V_i}$.

We use the percentile-rank metric defined in
Appendix~\ref{app:perturbation-construction}. For each edit family and direction, we first average scores across
subtrees from the same parent and then across parents, so each parent
contributes equally. Confidence intervals and paired model comparisons
follow Appendix~\ref{app:perturbation-construction}, using $1{,}000$
bootstrap resamples of parent expressions; all subtrees and variants
from a parent are kept together.
\begin{figure}[t]
    \centering
    \includegraphics[width=\columnwidth]
        {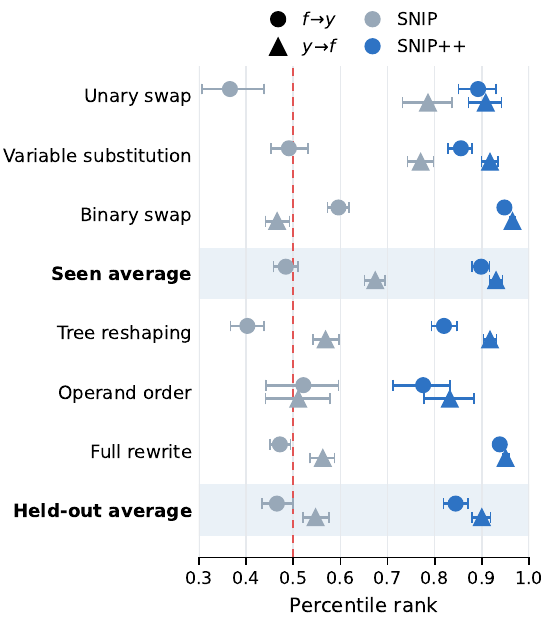}
    \caption{\textbf{Contextual-subtree discrimination.}
    Percentile rank of the correct subtree--behavior pairing across the six
    controlled edit families. The red dashed line marks chance ($.5$),
    and shaded rows report averages over the three
    training-represented and three held-out edit families. Error bars show
    $95\%$ bootstrap confidence intervals over parent expressions.}
    \label{fig:subtree-discrimination}
\end{figure}

\paragraph{Results.}

Figure~\ref{fig:subtree-discrimination} shows that \ourmodel{} substantially
improves contextual-subtree discrimination in both directions. Averaged
over the six edit families, its percentile rank reaches $.87$ for
$f\!\rightarrow\!y$ and $.92$ for $y\!\rightarrow\!f$, compared with $.47$
and $.61$ for SNIP, respectively.

The gains occur across all six families, including tree reshaping, operand
reordering, and full-subtree rewriting, which are held out during training.
Across the $12$ edit-family--direction settings, \ourmodel{} is significantly
better in all $12$, with no statistically indistinguishable settings or
reversals. These results show that the improved alignment is present within
contextualized local components, rather than arising only after pooling the
complete expression.

\section{Out-of-Domain Evaluation Details}
\label{app:ood-details}

Section~\ref{sec:ood-generalization} evaluates transfer to three external
symbolic-regression corpora: Feynman, Classic, and Strogatz. We additionally
report their pooled union.

\paragraph{Protocol.}

For each expression, every input feature is standardized before numerical
encoding. The same standardized inputs are used for SNIP and \ourmodel{}. We use the two protocols from
Section~\ref{sec:ood-generalization}: pooled cross-modal retrieval and
fine-grained discrimination against controlled symbolic variants. Both are
evaluated in the $f\!\rightarrow\!y$ and $y\!\rightarrow\!f$ directions.

For contextual-subtree evaluation, we pool the subtree span from its parent
encoding and match it to the numerical behavior computed by that subtree,
following Appendix~\ref{app:subtree-evaluation}. Scores are averaged first
within each parent expression and then across parents. Confidence intervals
are obtained from $1{,}000$ bootstrap resamples of parent expressions.

\begin{table}[t]
    \centering
    \small
    \begin{tabular}{lcc}
        \toprule
        \textbf{Evaluation set}
        & \textbf{Complete expressions}
        & \textbf{Subtree parents} \\
        \midrule
        Feynman  & 81  & 79  \\
        Classic  & 35  & 30  \\
        Strogatz & 13  & 12  \\
        \midrule
        Pooled union & 129 & 121 \\
        \bottomrule
    \end{tabular}
    \caption{\textbf{Out-of-domain evaluation pools.}
    Counts are reported after validity filtering.}
    \label{tab:ood-pools}
\end{table}

\paragraph{Contextual-subtree transfer.}

The local representations learned by \ourmodel{} also transfer beyond the
training generator. In fine-grained discrimination, its contextual-subtree
representations obtain an average percentile rank of $.68$, compared with
$.47$ for SNIP. In pooled contextual-subtree retrieval, \ourmodel{} reaches
$.78$, compared with $.48$ for SNIP. These results closely track the
corresponding complete-expression scores of $.69$ and $.79$, showing that
the transferred alignment is present both within local components and after
pooling the complete expression.

At the family level, transfer is strongest for variable substitutions and
operator swaps, but improvements also extend to several structural edit
families held out during training. 

\section{Training-Time Perturbation Construction}
\label{app:perturbations}
This appendix describes the controlled subtree perturbations used during
training in Section~\ref{sec:method-stloss}. The evaluation protocol in
Appendix~\ref{app:perturbation-construction} includes these same three edit
families, together with three additional families held out from training.
All evaluation expressions and individual edit instances are independently
generated and unseen during training.

\paragraph{Perturbation families.}

For each training expression $f_i$, we sample one non-trivial subtree
$s_{i0}$ and construct nearby symbolic variants using three edit families:

\begin{itemize}
    \item \textbf{Unary-operator swap.}
    One unary operator is replaced by another operator from the same
    compatible group:
    \[
        \{\sin,\cos,\tan\},\qquad
        \{(\cdot)^2,(\cdot)^3,\sqrt{\cdot}\},\qquad
        \{\exp,\log\}.
    \]
    This edit is unavailable when the subtree contains no supported unary
    operator.

    \item \textbf{Variable substitution.}
    One variable is replaced by a different variable already present in the
    subtree. All occurrences of the selected source variable are replaced
    consistently. This edit is unavailable when fewer than two distinct
    variables occur.

    \item \textbf{Binary-operator swap.}
    One occurrence of an operator in
    \[
        \{+,-,\times,\div\}
    \]
    is replaced by one of the other three operators, while its operands and
    the remainder of the subtree are unchanged.
\end{itemize}

Each edit changes one controlled symbolic choice while leaving the remaining
subtree structure unchanged.

\paragraph{Sampling and validity checks.}

We make two perturbation attempts for each sampled subtree. For each attempt,
we sample an applicable edit family, an eligible location, and a replacement.
An attempted edit is discarded if it does not produce a well-formed
expression or cannot be evaluated.

Let $s_{i1},\ldots,s_{iK_i}$ denote the retained variants, where
$K_i\leq 2$. The original subtree and all retained variants are evaluated on
the same input set $X_i$:
\begin{equation}
    y_{ik}=s_{ik}(X_i),
    \qquad
    k=0,\ldots,K_i.
    \label{eq:app-training-perturbation-behaviors}
\end{equation}

A candidate is retained only when every value in $y_{ik}$ is finite and its
output vector is not an exact duplicate of the original or of a previously
retained variant. Exact behavioral duplicates are removed by hashing the
output vectors.

\paragraph{Role in the training objective.}

In the initial contextual and standalone subtree objectives, the retained
perturbed behaviors act as hard numerical alternatives to the original
subtree. In the symmetric objective, each retained perturbation is also
encoded symbolically---both within its parent and independently---and paired
with its own numerical behavior. Finally, the perturbation is inserted into
the parent expression to provide an additional complete-expression training
pair.
\section{Implementation Details}
\label{app:implementation-details}

\paragraph{Data generation.}

Training expressions are generated online using the SNIP expression generator,
rather than drawn from a fixed corpus. The maximum input dimensionality is
$10$.

\paragraph{Architecture.}

All experiments use the SNIP backbone. The symbolic and
numerical encoders each contain eight Transformer layers with model dimension
$512$, feed-forward dimension $2048$, and $16$ attention heads. Both modalities
are projected to a shared $512$-dimensional latent space.

\ourmodel{} retains SNIP's numerical encoder, tokenizer, pooling operations,
and projection heads. Its only additional architectural parameters are the
four learned tree-decay scalars $\{\theta_c\}_{c=1}^{4}$ introduced by the
tree-structural positional encoding.
The symbolic
encoder, numerical encoder, numerical embedder, and projection bottleneck
contains approximately $67.2$M parameters for SNIP and $67.2$M plus four
scalars for \ourmodel{}.

\paragraph{Optimization.}

Each model is trained for $100$ epochs with $1{,}000$ optimizer updates per
epoch, for a total of $10^5$ updates. We train on four NVIDIA H100-80GB GPUs
with a global batch size of $64$. Optimization uses Adam with
$\beta_1=0.9$, $\beta_2=0.999$, and $\epsilon=10^{-8}$. The learning rate is
linearly warmed up for the first $10{,}000$ updates to a peak value of
$4\times10^{-5}$ and subsequently follows an inverse-square-root decay.
Weight decay is zero, gradients are clipped to a maximum norm of $0.5$, and
training uses automatic mixed precision with FP16.

\paragraph{Hyperparameter selection.}

We evaluate subtree-loss weights
$\lambda_{\mathrm{sub}}\in\{0,0.5,1.0\}$ and subtree temperatures
$\tau_s\in\{1.0,0.1\}$. The final configuration uses
$\lambda_{\mathrm{sub}}=1.0$ and $\tau_s=0.1$, selected using held-out
fine-grained alignment subject to no regression in global retrieval.

All variants are trained from scratch with training seed $0$.  Confidence intervals and
paired comparisons are obtained by bootstrapping target-expression clusters,
keeping all subtrees and controlled variants associated with a target together.



A complete \ourmodel{} training run takes approximately $21.4$ hours on four
NVIDIA H100-80GB GPUs, corresponding to roughly $86$ GPU-hours. Evaluation
jobs use one H100-80GB GPU, eight CPU cores, and $64$GB of system memory. The corresponding SNIP training run  takes $11.4$ hours on a single H100-80GB GPU.
All evaluations use frozen encoders.


\end{document}